\documentclass[10pt,twocolumn,letterpaper]{article}

\usepackage{cvpr}              

\usepackage{graphicx}
\usepackage{amsmath}
\usepackage{amssymb}
\usepackage{booktabs}

\usepackage{times}  
\usepackage{helvet}  
\usepackage{courier}  
\usepackage[hyphens]{url}  
\usepackage{natbib}  
\setcitestyle{numbers,square}
\usepackage{caption} 
\usepackage{xcolor}
\usepackage[ruled,vlined]{algorithm2e}
\usepackage{multirow}
\usepackage{colortbl} 
\usepackage{url}
\usepackage{amsfonts}
\usepackage{float}

\usepackage[pagebackref,breaklinks,colorlinks]{hyperref}

\usepackage[capitalize]{cleveref}
\crefname{section}{Sec.}{Secs.}
\Crefname{section}{Section}{Sections}
\Crefname{table}{Table}{Tables}
\crefname{table}{Tab.}{Tabs.}

\def\cvprPaperID{*****} 
\def\confName{CVPR}
\def\confYear{2023}

\title{SOGDet: Semantic-Occupancy Guided Multi-view 3D Object Detection}
\author{
Qiu Zhou\thanks{Joint first authors with equal contributions}\quad
Jinming Cao$^{1*}$\thanks{Corresponding Author}\quad
Hanchao Leng$^2$\quad
Yifang Yin$^3$\quad
Yu Kun$^2$\quad\\
Roger Zimmermann$^1$
\\
$^1$National University of Singapore, Singapore\qquad
$^2$Xiaomi Group, China\qquad
$^3$A*STAR, Singapore
}

\begin{document}

\maketitle

\begin{abstract}
In the field of autonomous driving, accurate and comprehensive perception of the 3D environment is crucial.
Bird's Eye View (BEV) based methods have emerged as a promising solution for 3D object detection using multi-view images as input.
However, existing 3D object detection methods often ignore the physical context in the environment, such as sidewalk and vegetation, resulting in sub-optimal performance. 
In this paper, we propose a novel approach called SOGDet (Semantic-Occupancy Guided Multi-view 3D Object Detection), that leverages a 3D semantic-occupancy branch to improve the accuracy of 3D object detection.  
In particular, the physical context modeled by semantic occupancy helps the detector to perceive the scenes in a more holistic view.
Our SOGDet is flexible to use and can be seamlessly integrated with most existing BEV-based methods.
To evaluate its effectiveness, we apply this approach to several state-of-the-art baselines and conduct extensive experiments on the exclusive nuScenes dataset.
Our results show that SOGDet consistently enhance the performance of three baseline methods in terms of nuScenes Detection Score (NDS) and mean Average Precision (mAP). 
This indicates that the combination of 3D object detection and 3D semantic occupancy leads to a more comprehensive perception of the 3D environment, thereby aiding build more robust autonomous driving systems.
The codes are available at: \href{https://github.com/zhouqiu/SOGDet}{https://github.com/zhouqiu/SOGDet}.
\end{abstract}

\section{Introduction}
Autonomous driving has become a burgeoning field for both research and industry, with a notable focus on achieving accurate and comprehensive perception of the 3D environment. 
Recently, Bird’s Eye View (BEV) based methods~\cite{huang2021BEVDet,huang2022BEVDet4d,li2022bevformer,li2022bevdepth} have attracted extensive attention in 3D object detection due to their effectiveness in reducing computational costs and footprints. 
The common paradigm is to take the multi-view images as inputs to detect objects, wherein the noticeable work BEVDet~\cite{huang2021BEVDet} serves as a strong baseline.
BEVDet first extracts image features from multi-view images using a typical backbone network such as ResNet~\cite{he2016deep}. 
The features are thereafter mapped to the BEV space with View Transformer~\cite{philion2020lift}, followed by a convolutional network and a target detection head. 
Inspired by BEVDet, following studies~\cite{li2022bevdepth,huang2022BEVDet4d,feng2022aedet} have integrated additional features into this framework, such as depth supervision~\cite{li2022bevdepth} and temporal modules~\cite{huang2022BEVDet4d}.

\begin{figure}[t!]
	\centering
	\centering
	\includegraphics[width=0.47\textwidth]{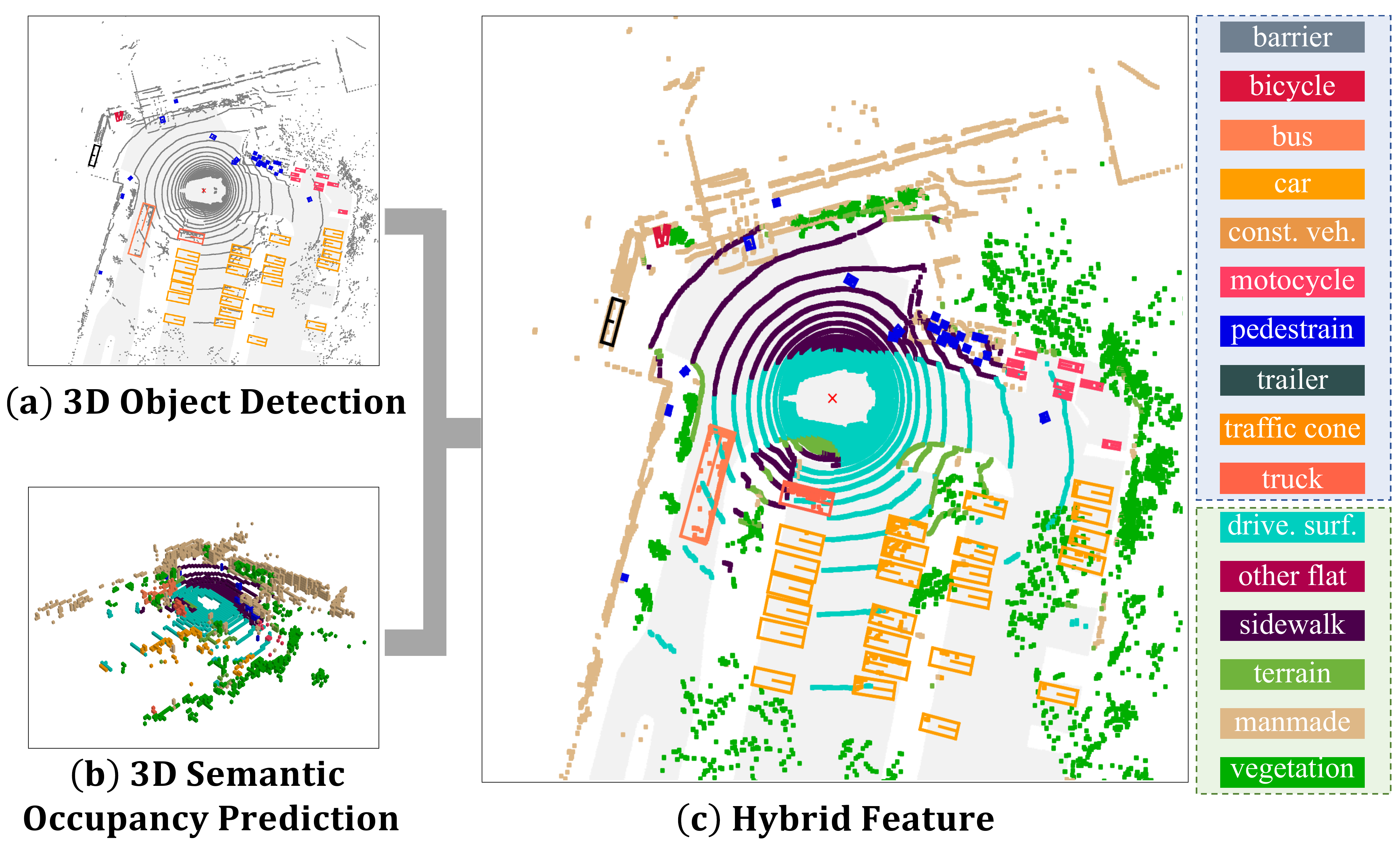}
	
	\caption{Illustration of 3D object detection and semantic occupancy prediction tasks. On the rightmost legend, the top 10 categories in the \textcolor{blue}{blue} box are shared for both tasks, and the bottom 6 categories in the \textcolor{green}{green} box are exclusively used by semantic occupancy prediction. (a) 3D object detection usually focuses on objects on roads, such as bicycles and cars. In contrast, 3D semantic occupancy prediction (b) concerns more about physical contexts (e.g., sidewalk and vegetation) in the environment. By combining these two (c), we can obtain a more comprehensive perception of the traffic conditions, such as pedestrians and bicycles mainly on the sidewalk and cars and buses co-appearing on drive surface.}
	\label{fig:teaser}
 \end{figure}

Despite the significant improvement in localizing and classifying specific objects such as cars, and pedestrians, most existing methods~\cite{huang2021BEVDet,huang2022BEVDet4d,li2022bevformer,li2022bevdepth} neglect the physical context in the environment.
These contexts, such as roads, pavements, and vegetation, though out of interest for detection, still offer important cues for perceiving the 3D scenes. For example, as shown in Figure~\ref{fig:teaser}, cars mostly appear in the drive surface rather than the sidewalk and vegetation.
To harness such important features for object detection, we notice a recent emerging task -- 3D semantic-occupancy prediction~\cite{huang2023tri,li2023voxformer,wei2023surroundocc,wang2023openoccupancy}, that voxelizes the given image and then performs semantic segmentation of the resulting voxels.
This task not only predicts the occupancy status but also identifies the objects within each occupied pixel, thereby enabling the comprehension of physical contexts.
As shown in Figure~\ref{fig:teaser}, 3D object detection and semantic occupancy prediction focuses on objects on roads and environmental contexts, respectively.
Combing these two leads to the hybrid features in Figure~\ref{fig:teaser}(c) that provide a more comprehensive description of the scene, such as the location and orientation of cars driving on the drive surface, and the presence of pedestrians on sidewalk or crossings.

Motivated by this important observation, we propose a novel approach called SOGDet, which stands for \textbf{S}emantic-\textbf{O}ccupancy \textbf{G}uided Multi-view 3D Object  \textbf{Det}ection. 
To the best of our knowledge, our method is the first of its kind to employ a 3D semantic-occupancy branch (OC) to enhance 3D object detection (OD). 
Specifically, we leverage a BEV representation of the scene to predict not only the pose and type of 3D objects (OD branch) but also the semantic class of the physical context (OC branch). 
SOGDet is a plug-and-play approach that can be seamlessly integrated with existing BEV-based methods~\cite{huang2021BEVDet, huang2022BEVDet4d, li2022bevdepth} for 3D object detection tasks in an end-to-end training manner. 
Moreover, to better facilitate the OD task, we extensively explore two labeling approaches for the OC branch, wherein the one predicts the \emph{binary occupancy} label only and the other involves the \emph{semantics} of each class. 
Based on these two approaches, we train two variants of SOGDet, namely SOGDet-BO and SOGDet-SE. Both variants significantly outperform the baseline method, demonstrating the effectiveness of our proposed method.

We conduct extensive experiments on the exclusive nuScenes~\cite{caesar2020nuscenes} dataset to evaluate the effectiveness of our proposed method. In particular, we apply SOGDet to several state-of-the-art backbone networks~\cite{he2016deep, liu2021swin,cao2021shapeconv} and compare it to various commonly used baseline methods~\cite{huang2022BEVDet4d, li2022bevdepth}.
Our experimental results demonstrate that SOGDet consistently improves the performance of all tested backbone networks and baseline methods on the 3D OD task in terms of nuScenes Detection Score (NDS) and mean Average Precision (mAP).
On the flip side, our OC approach surprisingly achieves comparable performance to state-of-the-art methods~\cite{huang2023tri}.
This finding represents another promising side product and is beyond our expectation, as our intention is to design a simple network and sheds little light on it.
The above results together highlight the effectiveness of the combination of 3D OD and OC in achieving comprehensive 3D environment understanding, and further enabling the development of robust autonomous driving systems.

\section{Related Work}\label{sec:related_work}

\subsection{3D Object Detection (OD)} constituents an indispensable component in autonomous driving~\cite{arnold2019survey,chen2017multi}. Prior monocular methods~\cite{ding2020learning,cai2020monocular,kumar2021groomed,reading2021categorical} predict 3D bounding boxes using single-view images. For example, D4LCN~\cite{ding2020learning} uses an estimated depth map to enhance image representation. Cai et al.~\cite{cai2020monocular} used object height prior to invert the 2D structured polygon into a 3D cuboid. However, due to the limitation of scarce data and single-view input, the model demonstrates difficulties in tackling more complex tasks~\cite{huang2021BEVDet}.
To overcome this problem, recent studies~\cite{huang2021BEVDet,huang2022BEVDet4d,li2022bevdepth} have been devoted to the development of large-scale benchmarks~\cite{caesar2020nuscenes,sun2020scalability} with multiple camera views.
For example, inspired by the success of FCOS~\cite{tian2019fcos} in 2D detection, FCOS3D~\cite{wang2021fcos3d} treats the 3D OD problem as 2D OD. 
Specifically, FCOS3D conducts perception solely in the image view and employs the strong spatial correlation of the targets' attributes with image appearance. 
Based on FCOS3D, PGD~\cite{wang2022probabilistic} presents using geometric relation graph to facilitate the targets' depth prediction.
Benefited from the DETR~\cite{carion2020end} method, some approaches have also explored the validity of Transformer, such as DETR3D~\cite{wang2022detr3d} and Graph-DETR3D~\cite{chen2022graph}.

Unlike the aforementioned methods, BEVDet~\cite{huang2021BEVDet} leverages the Lift-Splat-Shoot (LSS) based~\cite{philion2020lift} detector to perform 3D OD in multiple views. 
The framework is explicitly designed to encode features in the BEV space, making it scalable for multi-task learning, multi-sensor fusion, and temporal fusion~\cite{huang2022BEVDet4d}.
The framework is extensively studied by following work, such as BEVDepth~\cite{li2022bevdepth}, which enhances depth prediction by introducing a camera-aware depth network. Additionally, BEVDet4D~\cite{huang2022BEVDet4d} and BEVFormer~\cite{li2022bevformer} extend BEVDet from the temporal and spatiotemporal dimension, respectively.
Our proposed method also builds upon the BEVDet framework.
Specifically, we introduce the semantic occupancy branch to guide the prediction of object detectors, a paradigm that has not been studied by existing efforts.

\begin{figure*}[t!]
	\centering
	\centering
	\includegraphics[width=0.9\textwidth]{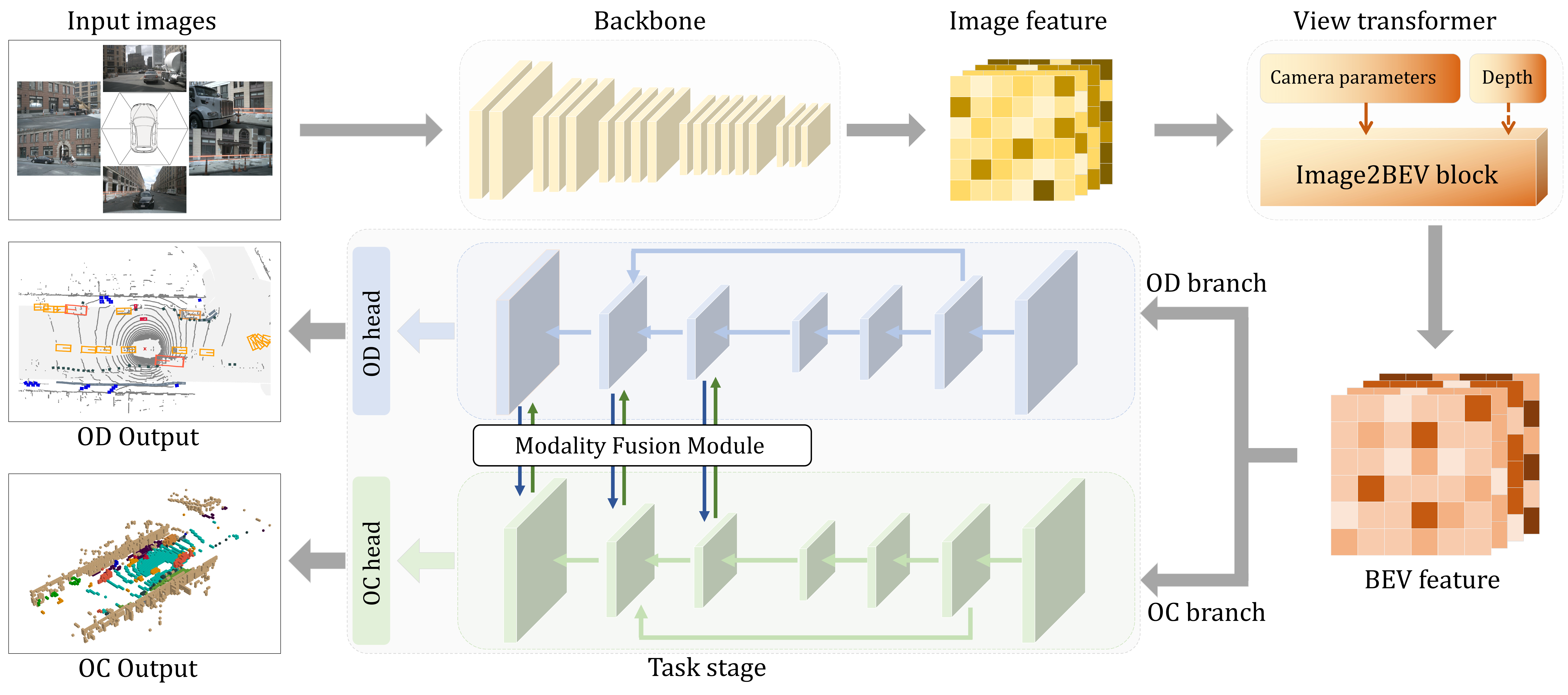}
 
	\caption{The overall network architecture. Our approach includes an \textcolor{yellow}{image backbone} to encode multi-view input images to the vision feature, a \textcolor{orange}{view transformer} to transform the vision feature into BEV feature, and a task stage comprising \textcolor{blue}{OD} and \textcolor{green}{OC} branches that respectively predict the OD and OC outputs in the same time.}
	\label{fig:Architecture}
 \end{figure*}
 
\noindent \subsection{3D Semantic Occupancy Prediction (OC)} has emerged as a popular task in the past two years~\cite{cao2022monoscene,huang2023tri,li2023voxformer,miao2023occdepth,wei2023surroundocc,wang2023openoccupancy}. It involves assigning an occupancy probability to each voxel in the 3D space. The task offers useful 3D representations for multi-shot scene reconstruction, as it ensures the consistency of multi-shot geometry and helps obscured parts to be recovered~\cite{shi2023grid}. 

The existing methods are relatively sparse in the literature. 
MonoScene~\cite{cao2022monoscene} is the pioneering work that uses monocular images to infer dense 3D voxelized semantic scenes. However, simply fusing multi-camera results with cross-camera post-processing often leads to sub-optimal results. VoxFormer~\cite{li2023voxformer} devises a two-stage framework to output the full 3D volumetric semantics from 2D images. The first stage uses a sparse collection of depth-estimated visible and occupied voxels, followed by a densification stage that generates dense 3D voxels from the sparse ones.
TPVFormer~\cite{huang2023tri} performs end-to-end training by using sparse LiDAR points as supervision, resulting in more accurate occupancy predictions.

It is worth noting that the methods discussed above are specifically designed to tackle the 3D OC task. However, our focus in this paper is to improve 3D OD by incorporating 3D semantic-occupancy as a supportive branch. 
As a result, we empirically resort to a simple network structure of OC to validate the key idea and do not adopt the standard OC labeling protocols following~\cite{huang2023tri}.

\section{Method}
\label{sec:method}

\subsection{Overall Architecture and Notations}

The overall architecture of our proposed method is illustrated in Figure~\ref{fig:Architecture}.
It is composed of three main components: an image backbone, a view transformer, and a task stage that predicts both OC and OD simultaneously. 
Specifically, the multi-view input images are first encoded by the image backbone, and then aggregated and transformed into the Bird-Eye-View (BEV) feature by the view transformer. With inherent camera parameters, the view transformer conducts depth-aware multi-view fusion and 4D temporal fusion simultaneously. Thereafter, the task stage generates both OC and OD features, which are interacted using a modality fusion module. We finally predict the OD and OC outputs using their respective features.

To ensure the clearance and consistency throughout our presentation, we first define the following notations following the order of data flow within our pipeline.
 
${I}$ represents an image group with same height and width from $N$ cameras with the same timestamp. 
$\boldsymbol{F_{img}} \in \mathbb{R}^{N \times C \times H \times W} $ represents feature map produced by the image backbone. ${H}$, ${W}$ and ${C}$ means the height, width and channels of the feature map, respectively. $\boldsymbol{F_d} \in \mathbb{R}^{N \times D \times H \times W} $ represents depth estimation of the multi-view image group ${I}$. $\boldsymbol{F_{bev}} \in \mathbb{R}^{C_{bev} \times X \times Y}$ represents BEV features extracted by the view transformer. The dimensions of the BEV plane are represented by $X \times Y$, and the number of channels in the BEV feature is $C_{bev}$, which is set to 128 following~\cite{huang2022BEVDet4d}. $\boldsymbol{F_{od}}$ and $\boldsymbol{F_{oc}}$ represent task-specific intermediate features of OD and OC branches in task stage. 

For the camera parameters, we combine the offset vector and rotation matrix to represent the translation $\boldsymbol{TR} \in \mathbb{R}^{4 \times 4}$  from source coordinate system to target coordinate system. For example, $\boldsymbol{{TR}_{cam}^{lid}}$ means a translation from camera coordinate system to lidar coordinate system. And $\boldsymbol{TR_{in}}$ represents the intrinsic parameters of all cameras.

For the output, the OD branch has two outputs: Bounding Box $\boldsymbol{B} \in \mathbb{R}^{M \times (3+3+2+2+1)} $ and Heatmap $\boldsymbol{H}$, where $M$ is the total number of bounding boxes and the second dimension of $\boldsymbol{B}$ represents location, scale, orientation, velocity and attribute. $\boldsymbol{Occ} \in \mathbb{R}^{O \times X \times Y \times Z} $ represents the OC branch output, which means that for the different grids from voxel grid $\boldsymbol{V} \in \mathbb{R}^{X \times Y \times Z} $, there are $O$ semantic labels in total. And we generate the occupancy voxel grid from point cloud $\boldsymbol{P} \in \mathbb{R}^{K \times 3} $ of $K$ points.

\subsection{Image Backbone}
The image backbone encodes the multi-view input images ${I}$ into the feature map $\boldsymbol{F_{img}}$. Following previous work~\cite{huang2021BEVDet,huang2022BEVDet4d}, we sequentially concatenate ResNet~\cite{he2016deep} and FPN~\cite{lin2017feature} as our image backbone to extract the image feature.
Moreover, we empirically found that using ShapeConv~\cite{cao2021shapeconv} instead of traditional convolutional layers in the image backbone leads to improved accuracy in the OD task without increasing model complexity during inference. 
In view of this, all ResNet-50 and -100 models in our method and baseine are replaced with ShapeConv for a fair comparison.
Detailed ablation studies on the performance obtained by ShapeConv can be found in the Supp.

\subsection{View Transformer}

The view transformer converts the image feature $\boldsymbol{F_{img}}$ to the BEV feature $\boldsymbol{F_{bev}}$. We implement this module with the combination of BEVDepth~\cite{li2022bevdepth} and BEVDet4D~\cite{huang2022BEVDet4d} for better performance, namely BEVDet4D-depth, which jointly conducts depth-aware multi-view fusion and 4D temporal fusion based on BEVDepth and BEVDet4D, respectively.

\subsubsection{Depth-aware Multi-view Fusion.} 
Following BEVDepth~\cite{li2022bevdepth}, the  $\boldsymbol{F_d}$ feature is estimated by a Depth Network based on image feature $\boldsymbol{F_{img}}$ and camera parameter $\boldsymbol{{TR}_{in}}$ by,
\begin{equation}
\boldsymbol{F_d} = DepthNet(\boldsymbol{F_{img}}, \boldsymbol{{TR}_{in}}).
\label{eqt:fd}
\end{equation}
Here, we use the notation $DepthNet(*,*)$ to refer to the sub-network introduced in~\cite{li2022bevdepth}, which is composed of a series of convolutional layers and MLPs.

Then the Lift-Solat-Shoot(LSS)~\cite{philion2020lift} is applied to  calculate BEV feature $\boldsymbol{F_{bev}}$ as follows,
\begin{equation}
\boldsymbol{F_{bev}} = LSS(\boldsymbol{F_{img}}, \boldsymbol{F_d}, \boldsymbol{{TR}_{cam}^{lid}} ),
\label{eqt:vtrans}
\end{equation}
where $LSS(*,*,*)$ is a depth-aware transformation based on ~\cite{philion2020lift} following~\cite{li2022bevdepth} which first lift the image feature $\boldsymbol{F_{img}}$ and its depth feature $\boldsymbol{F_d}$ into 3D lidar coordinate system by $\boldsymbol{{TR}_{cam}^{lid}}$, then splat 3D feature into 2D BEV plane to obtain $\boldsymbol{F_{bev}}$.

\subsubsection{4D Temporal Fusion.}
Let $\boldsymbol{F_{bev}^{curr}}$ and $\boldsymbol{F_{bev}^{adj}}$ represent the BEV feature in the current timestamp and an adjacent timestamp, respectively. We then apply a temporal fusion step following BEVDet4D~\cite{huang2022BEVDet4d} to aggregate $\boldsymbol{F_{bev}^{curr}}$ and $\boldsymbol{F_{bev}^{adj}}$ using Equation~\ref{eqt:tempo},
\begin{equation}
\boldsymbol{F_{bev}} = Concat[\boldsymbol{F_{bev}^{curr}},  \boldsymbol{F_{bev}^{adj}} ]
\label{eqt:tempo}
\end{equation}
where $Concat[*, *]$ represents the concatenation of two matrices along the channel dimension. 

\subsection{Task Stage}
The task stage consists of two branches that take the BEV feature $\boldsymbol{F_{bev}}$ as the input to obtain the Bounding Box $\boldsymbol{B}$ and Heatmap $\boldsymbol{H}$ outputs for OD branch and the Occpancy output $\boldsymbol{Occ}$ for OC branch, respectively.

On the one hand, the OD branch is our primary task branch, which performs a 10-class object detection on car, truck, etc. On the other hand, the OC branch is to facilitate object detection by generating a 3D geometrical voxel around the ego vehicle.

To refine the BEV feature $\boldsymbol{F_{bev}}$ in both branches, we first apply a 3-layers ResNet~\cite{he2016deep} to extract intermediate features $\boldsymbol{F_{od}}$ and $\boldsymbol{F_{oc}}$ in three different resolution, which are 1/2, 1/4, 1/8 of the height, width respectively. A pyramid network~\cite{lin2017feature} is then employed to upsample the features to the same size as the original one. For the OD branch, we use CenterPoint~\cite{yin2021center} to produce the final OD bounding box prediction heatmap $\boldsymbol{H}$ and bounding box $\boldsymbol{B}$ from $\boldsymbol{F_{od}}$. For the OC branch, a simple 3D-Conv Head~\cite{occupancy.org} is used to generate occupancy voxel grid $\boldsymbol{Occ}$ from $\boldsymbol{F_{oc}}$.


\subsubsection{Modality-fusion Module. } 
The modality-fusion module is essential in our method to perform interactions between the above two branches. We define $\mathbb{G}_{C \rightarrow D}$ to adapt the features from OC to OD, and vice versa with $\mathbb{G}_{D \rightarrow C}$.
We employ a weighted average operation parameterized by $\lambda$ to fuse features from different modalities and empirically set $\lambda=0.9$,
\begin{equation}
\left\{
\begin{aligned}
\boldsymbol{F_{od}} = (1-\lambda) \cdot \mathbb{G}_{C \rightarrow D} (\boldsymbol{F_{oc}}) + \lambda \cdot \boldsymbol{F_{od}}, \\
\boldsymbol{F_{oc}} = (1-\lambda) \cdot \mathbb{G}_{D \rightarrow C} (\boldsymbol{F_{od}}) + \lambda \cdot \boldsymbol{F_{oc}}.
\end{aligned}
\right.
\label{eqt:gate}
\end{equation}

\noindent Taking OC to OD as example, the Equation~\ref{eqt:gate} above shows that feature $\boldsymbol{F_{od}}$ in branch OD are $1-\lambda$ replaced by feature $\mathbb{G}_{C \rightarrow D}(\boldsymbol{F_{oc}})$ from branch OC. $\mathbb{G}_{C \rightarrow D}$ serves as a filter to reduce the modality gap between OD and OC. The operation takes effect when the BEV feature is upsampled in their own branches each time in the pyramid network~\cite{lin2017feature} mentioned above. We will demonstrate that this strategy can effectively enhance the information that is ignored by their original branch and thus fill the modality gap.

\begin{figure}[t!]
	\centering
	\centering
	\includegraphics[width=0.47\textwidth]{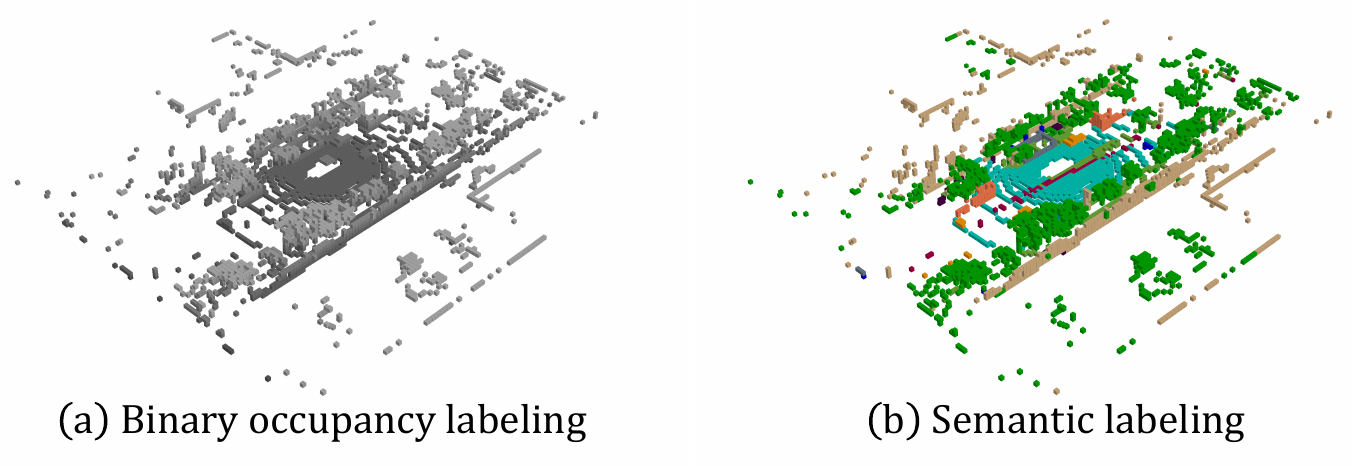}
 
	\caption{Illustration of the two types of labels.}
	\label{fig:cfl}
 \end{figure}

\subsection{Occupancy Label Generation} 
We leverage two types of supervision signals from the occupancy prediction for the OC branch. One is \emph{binary occupancy} label \textbf{BO}, whose supervision is binary with 0 and 1 representing empty and occupied voxels, respectively. The other is \emph{semantic} label \textbf{SE}, containing 16 semantic labels such as barrier, bicycle, etc. Figure~\ref{fig:cfl} illustrates the two types of label.

To generate the binary occupancy labels, we consider only the geometry features of each voxel and illustrate this procedure in Algorithm~\ref{alg:crslb}.
This approach is cost-friendly and require no extra manual annotations.

For semantic label, we observe that directly using the sparse semantic occupancy points as ground-truth labels leads to unstable training. Therefore, we follow TPVFormer~\cite{huang2023tri} to optimize the supervision voxel generation, where the voxels without semantic labels are masked and ignored. We detail this labeling process in the Supp.

\begin{algorithm}
\DontPrintSemicolon
\KwData{Point Cloud $\mathbf{P}$, Dimension Bound $X_{min}$, $X_{max}$, $Y_{min}$, $Y_{max}$, $Z_{min}$, $Z_{max}$, Resolution $R_X$, $R_Y$, $R_Z$}
\KwResult{Voxel Grid $\mathbf{V} $ }
\textcolor{blue}{/* Transform position of points into grid index */}\;
\For {$p \in \mathbf{P}$}{
    $p_X, p_Y, p_Z \gets p$\;
    \For {$axis \in \{X,Y,Z\} $}{
        \eIf{${axis}_{min}\leq  p_{axis}\leq {axis}_{max}$}{
            $p_{axis}  \gets \frac{(p_{axis} - {axis}_{min})}{R_{axis}}$\;
        }{
            
            $\mathbf{P}  \gets \mathbf{P}-\{p\}$ \textcolor{blue}{/* Delete out of bound */}\;
            break\;
        }
    }
}
\textcolor{blue}{/* Calculate the scale of output voxel grid */}\;
$X  \gets \frac{X_{max} - X_{min}}{R_X}$, $Y  \gets \frac{Y_{max} - Y_{min}}{R_Y}$, $Z  \gets \frac{Z_{max} - Z_{min}}{R_Z}$\;
build $\mathbf{V} \in \mathbb{R}^{X \times Y \times Z} $\;
\textcolor{blue}{/* Fill voxels */}\;
\For {$v \in \mathbf{V}$}{
    \textbf{if} $index(v) \in \mathbf{P}$  \textbf{then} $v  \gets 1$ \textbf{else} $v  \gets 0$\;
}

\caption{Binary occupancy label generation}\label{alg:crslb}
\end{algorithm}

\subsection{Training Objectives}

\subsubsection{Losses of OD Branch. }
We adopt the CenterPoint Head~\cite{yin2021center} to produce the final OD bounding box prediction, based on which a Gaussian focal loss~\cite{lin2017feature} and an L1 loss are jointly computed. In the following, we will sequentially elaborate these two loss functions.

Gaussian focal loss emphasizes more on the overall difference between predicted values and actual values across the entire plane. $\boldsymbol{H}$ denotes the heatmap output by the OD branch, which is a probability matrix recording the likelihood of each pixel belonging to any of the 10 classes. We then embed the real annotations into a 2D image with the same size as $\boldsymbol{H}$, forming the ground-truth heatmap $\boldsymbol{\widehat{H}}$, namely, a one-hot matrix. The Gaussian focal loss is then computed as,

\begin{equation}
{L}_{G} = -\lfloor \boldsymbol{\widehat{H}} \rfloor log(\boldsymbol{H}){(1-\boldsymbol{H})^\alpha}-{(1-\boldsymbol{\widehat{H}})}^{\gamma}log(1-\boldsymbol{H}){\boldsymbol{H}^\alpha},
\label{eqt:gauss}
\end{equation}
where $\lfloor * \rfloor$ denotes the floor operation, $\alpha=2.0$ and $\gamma=4.0$ are parameters of intensity following~\cite{lin2017focal}.

L1 loss is employed to optimize bounding box statistics, \emph{i.e.,} absolute distance location, scale, orientation, velocity and attribute, from a micro perspective. To this end, we estimate the L1 distance between predicted bounding box $\boldsymbol{B}$ and its ground-truth $\boldsymbol{\widehat{B}}$ as, 

\begin{equation}
{L}_{1} = \frac{1}{M} \cdot  \sum_m^M | \boldsymbol{B_m} - \boldsymbol{\widehat{B_m}} |.
\label{eqt:l1}
\end{equation}

In this way, the total loss of OD branch is shown as,
\begin{equation}
{L}_{OD}=L_G + \mu_{od}L_1,
\end{equation}
where $\mu_{od}$=0.25 is the weight coefficient of OD branch.

\subsubsection{Losses of OC Branch.}
We combine the cross entropy loss $L_{ce}$ with class weight and lov{\'a}sz-softmax loss~\cite{berman2018lovasz} $L_{lova}$ following~\cite{huang2023tri} in OC branch as follows,
\begin{equation}
{L}_{OC}=L_{lova} + \mu_{oc}L_{ce}
\label{eqt:lossoc}
\end{equation}
where $\mu_{oc}$=1 for SOGDet-SE and 6 for SOGDet-BO is the weight coefficient of OC branch. We set the same loss weight for all classes in SOGDet-SE and 1:2 for empty and occupied voxels in SOGDet-BO for loss weight in $L_{ce}$, respectively.

\subsubsection{Overall Objective.} 
Combined the above loss functions together, we can define our final objective as below,
\begin{equation}
L =  {L}_{OD}+ \omega{L}_{OC},
\label{eqt:lttl}
\end{equation}
where $\omega$ is the balancing factor between the OC and OD branches. We empirically set $\omega=10$ to maximize the effectiveness of our multi-task learning framework.

\begin{table}[]
\caption{\label{tab:val}Performance comparison on the nuScenes validation set. As indicated in~\cite{liu2021swin}, the complexity of Swin-Tiny and -Small are similar to those
of ResNet-50 and -101, respectively.
}
\centering
\scalebox{0.8}{
\begin{tabular}{c|c|c|c}
\hline
 Method                                    & Venue         & NDS(\%)↑      & mAP(\%)↑  \\ \hline\hline
PETR-Tiny                   & ECCV22        & 43.1          & 36.1      \\ 
BEVDet-Tiny             & arXiv22       & 39.2          & 31.2      \\  \hline
DETR3D-R50              & CoRL22        & 37.4          & 30.3      \\
Ego3RT-R50              & ECCV22        & 40.9          & 35.5      \\
BEVDet-R50             & arXiv22       & 37.9          & 29.8      \\
BEVDet4D-R50         & arXiv22       & 45.7          & 32.2      \\
BEVDepth-R50          & AAAI23        & 47.5          & 35.1      \\
AeDet-R50               & CVPR23        & 50.1          & 38.7      \\
\cellcolor[HTML]{EFEFEF}SOGDet-BO-R50         & \cellcolor[HTML]{EFEFEF}- & \cellcolor[HTML]{EFEFEF}50.2 & \cellcolor[HTML]{EFEFEF}38.2 \\ 
\cellcolor[HTML]{EFEFEF}SOGDet-SE-R50         & \cellcolor[HTML]{EFEFEF}- & \cellcolor[HTML]{EFEFEF}\textbf{50.6} & \cellcolor[HTML]{EFEFEF}\textbf{38.8} \\ \hline \hline

BEVerse-Small          & arXiv22       & 49.5          & 35.2      \\ \hline
PETR-R101                 & ECCV22        & 42.1          & 35.7      \\
UVTR-R101            & NeurIPS2022   & 48.3          & 37.9       \\
PolarDETR-T-R101          & arXiv22       & 48.8          & 38.3      \\
BEVFormer-R101          & ECCV22        & 51.7          & 41.6      \\
BEVDepth-R101            & AAAI23        & 53.5          & 41.2      \\
PolarFormer-R101   & AAAI23        & 52.8          & 43.2      \\
AeDet-R101                & CVPR23        & 56.1          & 44.9      \\
\cellcolor[HTML]{EFEFEF}SOGDet-BO-R101         & \cellcolor[HTML]{EFEFEF}- & \cellcolor[HTML]{EFEFEF}55.4         & \cellcolor[HTML]{EFEFEF}43.9      \\
\cellcolor[HTML]{EFEFEF}SOGDet-SE-R101         & \cellcolor[HTML]{EFEFEF}- & \cellcolor[HTML]{EFEFEF}\textbf{56.6}          & \cellcolor[HTML]{EFEFEF}\textbf{45.8}          \\ \hline
\end{tabular}
}

\end{table}

\section{Experiments}

\label{sec:experiments}

\begin{table*}[ht!]
\centering
\caption{\label{tab:test}Performance comparison on the nuScenes test set.
}
\scalebox{0.85}{
\begin{tabular}{c|c|c|c|ccccc}
\hline
Method                     & Venue   & NDS(\%)↑ & mAP(\%)↑ & mATE↓ & mASE↓ & mAOE↓ & mAVE↓ & mAAE↓ \\ \hline \hline
FCOS3D~\cite{wang2021fcos3d}                     & ICCV21  & 42.8     & 35.8     & 0.690  & 0.249 & 0.452 & 1.434 & 0.124 \\
DD3D~\cite{park2021pseudo}                       & ICCV21  & 47.7     & 41.8     & 0.572 & 0.249 & \textbf{0.368} & 1.014 & 0.124 \\
PGD~\cite{wang2022probabilistic}                       & CoRL22  & 44.8     & 38.6     & 0.626 & 0.245 & 0.451 & 1.509 & 0.127 \\

BEVDet~\cite{huang2021BEVDet}                    & arXiv22 & 48.2     & 42.2     & 0.529 & \textbf{0.236} & 0.395 & 0.979 & 0.152 \\

BEVFormer~\cite{li2022bevformer}                 & ECCV22  & 53.5     & 44.5     & 0.631 & 0.257 & 0.405 & 0.435 & 0.143 \\
DETR3D~\cite{wang2022detr3d}                     & CoRL22  & 47.9     & 41.2     & 0.641 & 0.255 & 0.394 & 0.845 & 0.133 \\
Ego3RT~\cite{lu2022learning}                      & ECCV22  & 47.3     & 42.5     & 0.549 & 0.264 & 0.433 & 1.014 & 0.145 \\
PETR~\cite{liu2022petr}                      & ECCV22  & 50.4     & 44.1     & 0.593 & 0.249 & 0.383 & 0.808 & 0.132 \\
CMT-C~\cite{yan2023cross} & ICCV23  & 48.1  & 42.9  & 0.616 & 0.248 & 0.415 & 0.904 & 0.147 \\
PETRv2~\cite{liu2022petrv2} & ICCV23 & 55.3 & 45.6 & 0.601 & 0.249 & 0.391 & 0.382 & \textbf{0.123} \\
X3KD~\cite{klingner2023x3kd} & CVPR23 & 56.1  & 45.6 & 0.506 & 0.253 & 0.414  & 0.366 & 0.131  \\

\rowcolor[HTML]{EFEFEF} 
SOGDet-BO                       & -       & 57.8 & 47.1 & 0.482 & 0.248 & 0.390 & \textbf{0.329} & 0.125 \\  
\rowcolor[HTML]{EFEFEF} 

SOGDet-SE                       & -       & \textbf{58.1} & \textbf{47.4} & \textbf{0.471} & 0.246 & 0.389 & 0.330 & 0.128 \\ \hline 
\end{tabular}
}

\end{table*}

\begin{table*}[h!]
\centering
\caption{\label{tab:tpvf} Comparison with the State-of-the-Art OC method on the nuScenes val set.}
\scalebox{0.75}{
\begin{tabular}{c|c|c|cccccccccccccccc}
\hline
                                  &                                  &                                      & \multicolumn{16}{c}{category-wise IoU   (\%)↑}                                                       \\ \cline{4-19} 
\multirow{-2}{*}{\textbf{Method}} & \multirow{-2}{*}{\textbf{Venue}} & \multirow{-2}{*}{\textbf{mIoU(\%)↑}} & $\raisebox{-.2\height}{\includegraphics{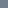}}$ & $\raisebox{-.2\height}{\includegraphics{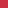}}$& $\raisebox{-.2\height}{\includegraphics{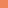}}$ &$\raisebox{-.2\height}{\includegraphics{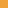}}$& $\raisebox{-.2\height}{\includegraphics{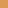}}$  &$\raisebox{-.2\height}{\includegraphics{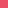}}$ &$\raisebox{-.2\height}{\includegraphics{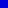}}$  &$\raisebox{-.2\height}{\includegraphics{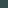}}$&$\raisebox{-.2\height}{\includegraphics{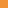}}$&$\raisebox{-.2\height}{\includegraphics{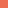}}$ &$\raisebox{-.2\height}{\includegraphics{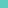}}$&$\raisebox{-.2\height}{\includegraphics{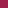}}$&$\raisebox{-.2\height}{\includegraphics{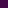}}$ & $\raisebox{-.2\height}{\includegraphics{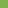}}$  &$\raisebox{-.2\height}{\includegraphics{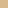}}$ & $\raisebox{-.2\height}{\includegraphics{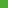}}$     \\ \hline\hline
TPVFormer                        & CVPR23                           & \textbf{59.3}                        & \textbf{64.9} & 27.0            & \textbf{83.0} & \textbf{82.8} & 38.3          & 27.4        & \textbf{44.9} & 24.0            & 55.4          & \textbf{73.6} & \textbf{91.7} & \textbf{60.7} & \textbf{59.8} & \textbf{61.1} & \textbf{78.2} & 76.5          \\

\rowcolor[HTML]{EFEFEF} 
SOGDet-SE                           & -                                & 58.6                                 & 57.8          & \textbf{30.7} & 74.9        & 74.7          & \textbf{43.7} & \textbf{42.0} & 44.5          & \textbf{32.7} & \textbf{62.6} & 63.9          & 85.9          & 54.3          & 54.6          & 58.9          & 76.9          & \textbf{80.2} \\ \hline
\end{tabular}
}

\end{table*} 

\subsection{Experimental Setup}

\subsubsection{Dataset and metrics.}
We conducted extensive experiments on the nuScenes~\cite{caesar2020nuscenes} and Panoptic nuScenes~\cite{fong2022panoptic} dataset, which is currently the exclusive benchmark for both 3D object detection and occupancy prediction. 
Following the standard practice~\cite{huang2021BEVDet,feng2022aedet}, we used the official splits of this dataset: 700 and 150 scenes respectively for training and validation, and the remaining 150 for testing.

For OD task, we reported nuScenes Detection Score (NDS), mean Average Precision (mAP), mean Average Translation Error (mATE), mean Average Scale Error (mASE), mean Average Orientation Error (mAOE), mean Average Velocity Error (mAVE), and mean Average Attribute Error (mAAE). Among them, NDS and mAP are the more representative ones.

For OC task, we designed two types of occupancy labeling approaches.
For the binary occupancy labeling approach, to the best of our knowledge, we are the first to employ such labeling approach in the literature, 
we performed the qualitative experiments.
For the semantic labeling one, we maintained a consistent experimental protocol with the state-of-the-art method TPVFormer\cite{huang2023tri}. Accordingly, we report the mean Intersection over Union (mIoU) of all semantic categories.

\subsubsection{Implementation details.}
To demonstrate the effectiveness and generalization capabilities of SOGDet, we used several popular architectures~\cite{li2022bevdepth,huang2022BEVDet4d}. To ensure that any improvements were solely due to our SOGDet, we kept most experimental settings, such as backbone and batch size untouched, and added only the OC branch. Unless otherwise noted, our baseline model is BEVDet4D-depth, which is a fusion of two recent multi-view 3D object detectors, BEVDepth~\cite{li2022bevdepth} and BEVDet4D~\cite{huang2022BEVDet4d} as described in Section~\ref{sec:method}. We followed the experimental protocol of AEDet~\cite{feng2022aedet} and training on eight 80G A100 GPUs with a mini-batch size of 8, for a total batch size of 64, and trained the model for 24 epochs with CBGS~\cite{zhu2019class} using AdamW as the optimizer with a learning rate of 2e-4. 

\subsection{Comparison with State-of-the-Art}

We evaluated the performance of our SOGDet model against other state-of-the-art multi-view 3D object detectors on the nuScenes validation and test sets. 

Table~\ref{tab:val} reports the results for the validation set using Swin-Tiny, -Small, ResNet-50 and -101 backbones (detailed results can be found at the Supp.). As shown in the table, our method achieves highly favorable model performance, with NDS scores of 50.2\% and 55.4\% for SOGDet-BO and 50.6\% and 56.6\% for SOGDet-SE on ResNet-50 and -101, respectively. These results surpass current state-of-the-art multi-view 3D object detectors with a large margin, including BEVDepth~\cite{li2022bevdepth} (3.1\% improvement in NDS at both ResNet-50 and -100) and AEDet~\cite{feng2022aedet} (0.5\% improvement in NDS at both ResNet-50 and -100).

In Table~\ref{tab:test}, we present the results obtained by SOGDet with the ResNet-101 backbone on the nuScenes test set, where we report the performance of state-of-the-art methods that use the same backbone network for a fair comparison. We follow the same training strategy of existing approaches~\cite{li2022bevdepth,feng2022aedet} that utilize both the training and validation sets to retrain the networks and without any test-time augmentation. SOGDet shows improved performance in multi-view 3D OD task with 58.1\% NDS and 47.4\% mAP, further verifying the effectiveness of our proposed approach.

\begin{figure*}[ht!]

	\centering
	\centering
	\includegraphics[width=0.98\textwidth]{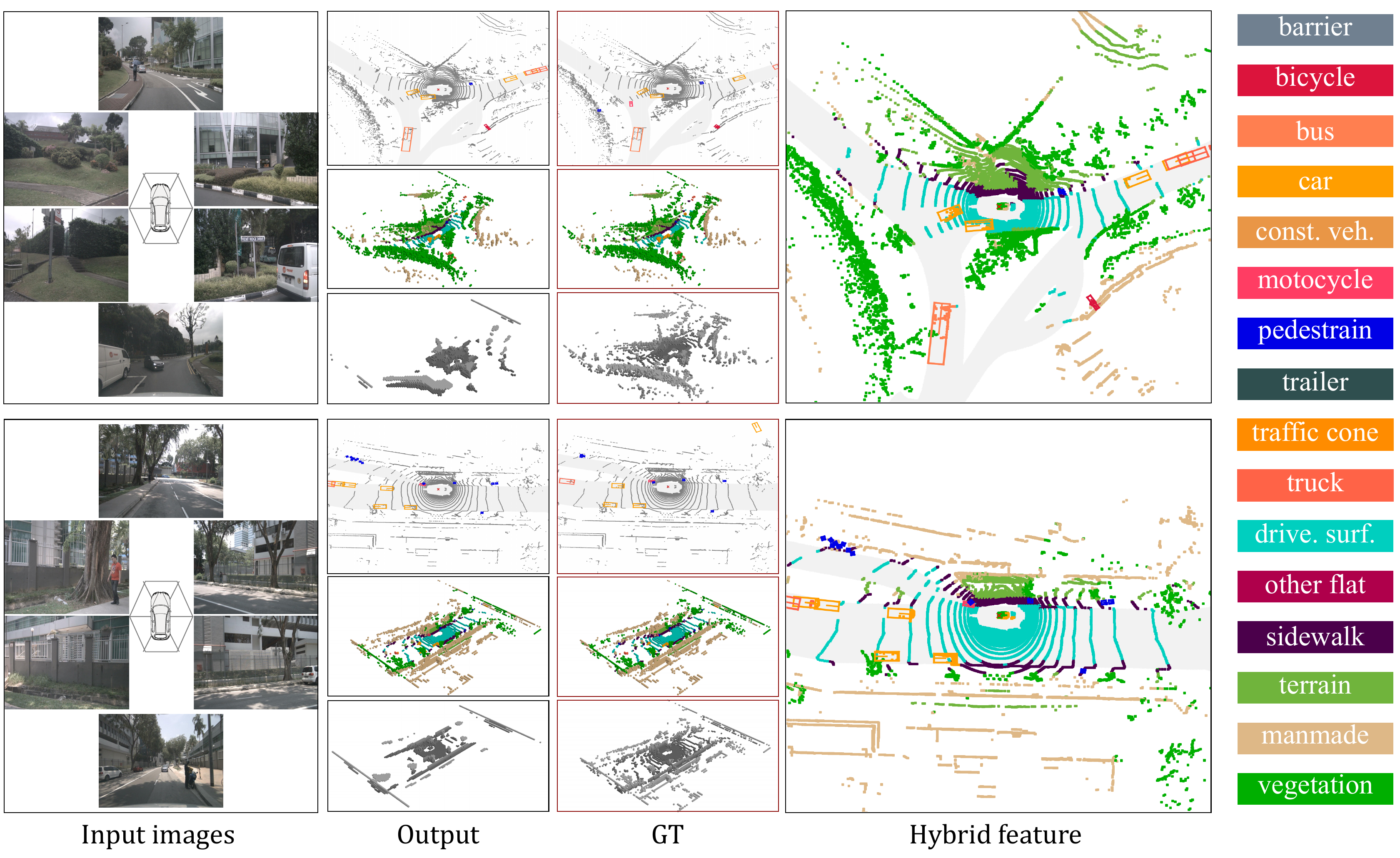}
	\caption{Visualization for the OD and OC branches of SOGDet. The input consists of six multi-view images. For both the output and the GT (red box) column, from top to bottom, we sequentially show the predictions of SOGDet-SE for OD, SOGDet-SE for OC and SOGDet-BO for OC.
 The Hybrid feature is blended from OD and OC branch predictions of SOGDet-SE.
 }
	\label{fig:vis}
 \end{figure*}

\subsection{Visualization}

Figure~\ref{fig:vis} illustrates qualitative results of our approach on the nuScenes~\cite{caesar2020nuscenes} dataset using ResNet-50 as the backbone for both OD and the OC branch, more results can be found in the Supp. 
Pertaining to the object detection task, we focus only on occupied voxels, and therefore, locations marked as ``empty'' are not shown. The hybrid features reveal strong correlations between the physical structures and the location of the detected objects, such as vehicles, bicycles, and pedestrians. For example, vehicles are typically detected in drive surface, while bicycles and pedestrians are often detected on sidewalk.
These findings are consistent with the observations and motivations of our paper and demonstrate that the integration of the two branches can lead to a better perception and understanding of the real world.

\subsection{Ablation Study}

\subsubsection{Comparison with the State-of-the-Art OC method} 
To further evaluate the effectiveness of our approach, we compared our method with respect to semantic categories of multi-view image input with TPVFormer~\cite{huang2023tri} and presented the results in Table~\ref{tab:tpvf}. Backbones from both methods take equivalent complexities.

The primary goal of our work is to enhance the 3D OD by integrating 3D OC. 
Despite its simpleness, our results, as shown in Table~\ref{tab:tpvf}, demonstrate that our SOGDet are comparable to TPVFormer, a state-of-the-art method specifically designed for the OC task. 
Moreover, our method even outperforms this baseline in certain categories such as bicycles, vegetation, and others.
This result indicates that the combination of the two branches can bring benefits for the OC branch as well, serving as another byproduct.

\begin{table}[h!]
\caption{\label{tab:baseline} Performance comparison with different baselines.}
\scalebox{0.77}{
\begin{tabular}{c|c|c|cc}
\hline
Backbone               & Architecture                     & Method                          & mAP(\%)↑                     & NDS(\%)↑                     \\ \hline
                        &                                  & Baseline                        & 31.2                         & 39.2                         \\
                        & \multirow{-2}{*}{BEVDet}       & \cellcolor[HTML]{EFEFEF}SOGDet-SE & \cellcolor[HTML]{EFEFEF}32.9 & \cellcolor[HTML]{EFEFEF}41.5 \\ \cline{2-5} 
                       &                                  & Baseline                        & 33.8                         & 47.6                         \\
\multirow{-4}{*}{Tiny} & \multirow{-2}{*}{BEVDet4D}       & \cellcolor[HTML]{EFEFEF}SOGDet-SE & \cellcolor[HTML]{EFEFEF}34.6 & \cellcolor[HTML]{EFEFEF}48.7 \\ \hline
                       &                                  & Baseline                        & 35.1                         & 47.5                        \\
                       & \multirow{-2}{*}{BEVDepth}       & \cellcolor[HTML]{EFEFEF}SOGDet-SE & \cellcolor[HTML]{EFEFEF}37.2 & \cellcolor[HTML]{EFEFEF}48.3 \\ \cline{2-5} 
                       &                                  & Baseline                        & 37.0                         & 49.0                         \\
\multirow{-4}{*}{R50}  & \multirow{-2}{*}{BEVDet4D-depth} & \cellcolor[HTML]{EFEFEF}SOGDet-SE & \cellcolor[HTML]{EFEFEF}38.8 & \cellcolor[HTML]{EFEFEF}50.6 \\ \hline
\end{tabular}
}

\end{table}

\subsubsection{Different baseline architecture}
Our proposed SOGDet is a flexible method that can be seamlessly integrated into most BEV-based multi-view object detection architectures. In order to evaluate the generalization capabilities of our method, we tested its effectiveness on several representative baseline architectures, namely BEVDet~\cite{huang2021BEVDet}, BEVDet4D~\cite{huang2022BEVDet4d}, BEVDepth~\cite{li2022bevdepth}, and BEVDet4D-depth, using the nuScenes validation set. The results, in Table~\ref{tab:baseline} show that SOGDet consistently surpasses these baselines under various settings. This result demonstrates the validity of our method to generalize to different model architectures.

\begin{figure}[h!]
	\centering
	\centering
	\includegraphics[width=0.43\textwidth]{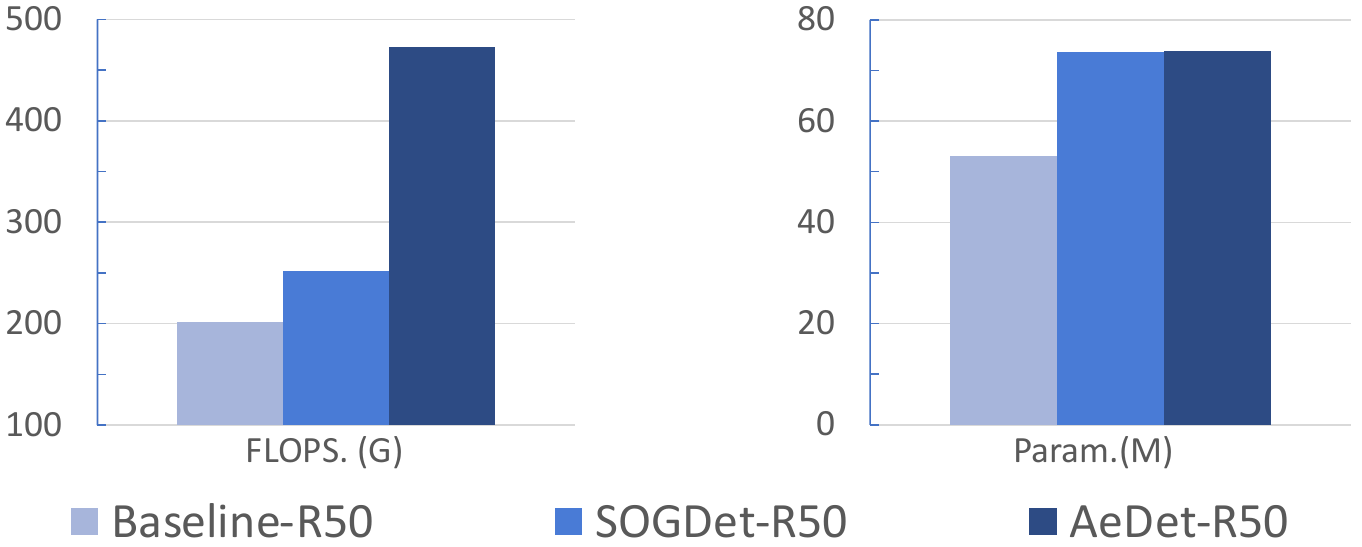}
 
	\caption{Parameter count (Param.) and floating-point operations (FLOPs).}
	\label{fig:flops}
 \end{figure}




\subsubsection{Complexity Analysis}

The efficiency concern is highly significant under resource-constrained environments. To study the effectiveness of our method pertaining to this aspect, we estimate metrics including floating point operations (FLOPs.) and parameter count (Param.), and show the results in Figure ~\ref{fig:flops}. It can be observed that compared with the SoTA method AeDet~\cite{feng2022aedet}, our SOGDet is more efficient especially on the more important metric FLOPs, i.e., 252G v.s. 473G. On the other hand, SOGDet outperforms AeDet by 0.5\% in terms of NDS. This indicates that our method achieves a better trade-off between efficiency and model performance.


Further ablation experiments such as performance of the two OC labeling approaches, different backbones, and hyperparameters, etc., can be found at the Supp.
\

\section{Conclusion and future work}
\label{sec:conclusion}
The Bird’s Eye View (BEV) based method has shown great promise in achieving accurate 3D object detection using multi-view images. However, most existing BEV-based methods unexpectedly ignore the physical contexts in the environment, which is critical to the perception of 3D scenes. In this paper, we propose the SOGDet approach to incorporate such context using a 3D semantic occupancy approach.
In particular, our SOGDet predicts not only the pose and type of each 3D object, but also the semantic classes of the physical contexts for finer-grained detection.
Extensive experimental results on the nuScenes dataset demonstrate that our SOGDet consistently improves the model performance of several popular backbone networks and baseline methods. 

In future work, we plan to explore the application of SOGDet with more auxiliary data inputs, such as lidar and radar, to further help the 3D object detection. Additionally, we believe that integrating 3D semantic-occupancy prediction into other autonomous driving tasks beyond 3D object detection, such as path planning and decision-making, may contribute a promising avenue for future research.


\end{document}